\pdfoutput=1
\documentclass[11pt]{article}

\usepackage[preprint]{acl-style-files/latex/acl}
\usepackage{times}
\usepackage{latexsym}
\usepackage[T1]{fontenc}
\usepackage[utf8]{inputenc}
\usepackage{microtype}
\usepackage{inconsolata}
\usepackage{graphicx}

\title{Searching for the Most Human-like Emergent Language}

\author{Brendon Boldt \and David Mortensen \\
  Language Technologies Institute \\
  Carnegie Mellon University \\
  Pittsburgh, PA, USA \\
  \texttt{\{bboldt,dmortens\}@cs.cmu.edu}}

\usepackage{hyperref}
\usepackage{xurl}
\usepackage[table]{xcolor}
\usepackage{latexsym,amsmath,amssymb}
\usepackage{booktabs}
\usepackage{enumitem}
\usepackage[nameinlink,capitalize]{cleveref}
\usepackage{pgfplots}
\usepackage{tikz}
\usetikzlibrary{shapes,arrows,shadows.blur,calc,positioning,fit,backgrounds}
\usepackage{algorithm,listings}

\newcommand\smallish{\fontsize{10pt}{10pt}\selectfont}

\newcounter{comment}

\everymath=\expandafter{\the\everymath\displaystyle}

\makeatletter\@ifpackageloaded{underscore}{}{\usepackage[strings]{underscore}}\makeatother

\begin{document}
\maketitle

\begin{abstract}
In this paper, we design a signalling game-based emergent communication environment to generate state-of-the-art emergent languages in terms of similarity to human language.
This is done with hyperparameter optimization, using XferBench as the objective function.
XferBench quantifies the statistical similarity of emergent language to human language by measuring its suitability for deep transfer learning to human language.
Additionally, we demonstrate the predictive power of entropy on the transfer learning performance of emergent language as well as corroborate previous results on the entropy-minimization properties of emergent communication systems.
Finally, we report generalizations regarding what hyperparameters produce more realistic emergent languages, that is, ones which transfer better to human language.
\end{abstract}

\section{Introduction}

Emergent language has tremendous potential to generate realistic human language data for deep learning methods without the need to collect data directly (or indirectly) from humans \citep{boldt2024review}.
This stems from the fact that emergent language aims to replicate the communicative pressures that drive the development of human language and are hypothesized to explain various patterns observed in linguistics \citep{sep-linguistics}.
Yet little work has been done to date designing emergent communication systems to generate languages with high statistical similarity to human languages.
Such languages could better serve as synthetic human language data for pretraining and evaluating NLP models.
Thus, in this paper, we generate emergent languages with a signalling game that have a high degree of similarity to human languages, demonstrating state-of-the-art performance on emergent-to-human language deep transfer learning.
Specifically, we use Bayesian hyperparameter search to optimize a signalling game on the XferBench benchmark \citep{boldt-mortensen-2024-xferbench}.

\begin{figure}
  \centering
  \input{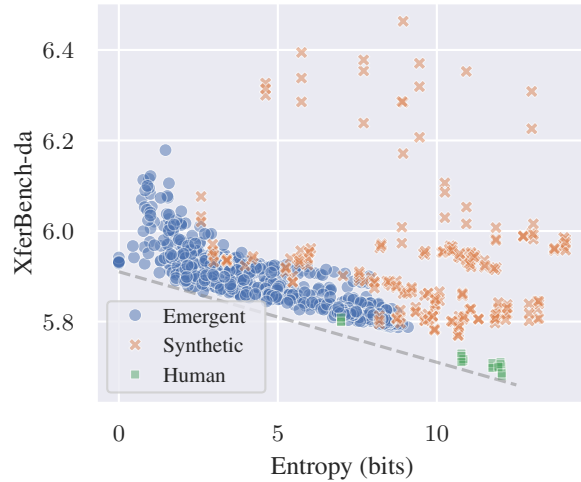}
  \caption{Hyperparameter search shows that emergent and human languages tend towards the Pareto frontier of minimizing entropy and minimizing XferBench score (lower is better) while non-emergent synthetic languages less reliably follow this trend.  Dashed gray line represents a lower bound on entropy versus XferBench score.}%
  \label{fig:ent-xb}
\end{figure}

Producing emergent languages which are more realistic (i.e., similar to human language) is one of the core goals of the field as a whole since the utility of emergent language is often predicated on its resemblance to human language \citep{boldt2024review}.
This paper takes a direct, principled approach to this goal by finding hyperparameters which maximize an emergent languages similarity to human language from a statistical perspective.
Such an approach is in stark contrast to a more arbitrary approach to selecting hyperparameters which is common in the methods of emergent communication.
For example, vocabulary sizes in emergent languages are often very small (only one of eight emergent language environments surveyed in \citet{elcc} exceeds a vocabulary size of $70$) while our research suggests that the optimal vocabulary size is in the $1$k to $10$k range.
Increasing vocabulary sizes, then, not only improves transfer learning performance but also makes it possible for emergent languages to replicate the long-tailed, Zipfian word distribution that is characteristic of human language \citep{zipf1949least,piantadosi2014zipf}, for example.
We produce a handful of such hyperparameter recommendations based on our empirical evaluations.

Beyond these recommendations,
  our experiments also confirm a significant relationship between transfer learning performance and corpus entropy.
Not only does it appear that the entropy of a corpus determines a lower bound on XferBench score (lower is better) but that emergent languages minimize entropy with respect to a given XferBench score in a way that procedurally generated (i.e., non-emergent, synthetic) languages do not (see \cref{fig:ent-xb}).
Such minimization is, significantly, an \emph{emergent} phenomenon as neither entropy nor transfer learning performance are directly involved in the optimization of the emergent communication system (and neither entropy nor XferBench incorporate each other).
This observation is significant in two regards:
  First, it suggests that transfer learning and, consequently, statistical similarity to human language can be (partially) explained with information theory.
  Second, it aligns closely with prior work that finds that emergent communication minimizes entropy with respect to task success within the environment \citep{kharitonov2020entmin,chaabouni2022emergent}.

We discuss related work in \cref{sec:related-work}.
Methods are discussed in \cref{sec:methods}, and the experiments are presented in \cref{sec:experiments}.
An analysis of the results is performed in \cref{sec:analysis} with discussion and conclusion in \cref{sec:discussion,sec:conclusion}.

\paragraph{Contributions}

We (1) introduce emergent communication environments which produce the most human language-like emergent languages to date, as shown by state-of-the-art performance on a deep transfer learning task using the XferBench benchmark;
(2) provide concrete recommendations on better hyperparameter settings for emergent communication experiments so as to make them more statistically similar to human language; and
(3) provide evidence that entropy minimization is a general property of emergent communication systems, finding that it is minimized with respect to transfer learning performance.

\section{Related Work}%
\label{sec:related-work}
At a high level, emergent communication (also called \emph{emergent language}) combines natural language processing, deep multi-agent reinforcement learning, and linguistics to study how natural language-like communication systems evolve or emerge from scratch.
One of the primary aims of this field is to discover what features of human language (e.g., compositionality) emerge from the environment and learning dynamics of the agents.
For a general overview of deep learning-based emergent communication research, see \citet{lazaridou2020emergentmultiagentcommunicationdeep}.
For the most part, this paper does not have any directly related work as optimizing emergent languages themselves across multiple game instances is relatively unexplored.
Below we present some particular facets of this paper which overlap with prior work.

This paper shares the goal of producing emergent language corpora that are suitable for transfer learning to human languages with \citet{yao2022linking},
  although \citet{yao2022linking} do not optimize the emergent languages directly and focus on validating the \emph{corpus transfer} technique (i.e., the basis of XferBench).
\citet{boldt2023mathmodel}, similarly to this paper, investigate the effect of hyperparameters on emergent communication, although their study focuses primarily on mathematically analyzing and explaining the effects rather than optimizing the emergent language for an evaluation metric.
Finally, this paper scales up emergent communication game hyperparameters in a way that overlaps with \citet{chaabouni2022emergent}, although the latter focuses on addressing the practical challenges of scaling up certain facets of the signalling game (e.g., number of agents) rather than directly optimizing for a particular objective.

The task of generating emergent languages for pretraining NLP models falls within the broad category data augmentation with synthetic data  but differs from most other approaches due emergent language's unique nature as an \emph{emergent} phenomenon.
First, emergent language differs from procedurally generating data from rules because emergent techniques preclude stipulating the exact process for generating the data; expert knowledge is incorporated into designing the system which generates the data, not generating the data itself.
On the other hand, emergent language differs from using pretrained language models to generate synthetic data since emergent communication is derived from scratch, again precluding any (pre)training on human language data.

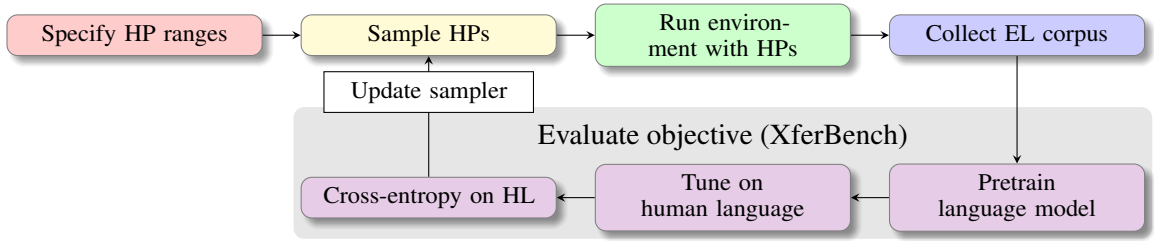
\begin{figure*}
  \centering
  \tikzstyle{blockStyle}=[
  draw=black!50,
  fill=violet!20,
  text width=3.1cm,
  text centered,
  blur shadow={shadow blur steps=5},
  rounded corners,
  font=\small,
]

\pgfdeclarelayer{background}
\pgfsetlayers{background,main}

\begin{tikzpicture}[
  node distance=5mm,
]

  \begin{scope}[local bounding box=bb]
    \node (beta) [blockStyle,fill=violet!20] {Pretrain\\language model};
    \node (gamma) [blockStyle,fill=violet!20,left=of beta] {Tune on\\human language};
    \node (delta) [blockStyle,fill=violet!20,left=of gamma] {Cross-entropy on HL};
    \node (title) at ($(beta)!0.5!(delta)$) [yshift=8mm] {Evaluate objective (XferBench)};
    \draw [-stealth] (beta) to (gamma);
    \draw [-stealth] (gamma) to (delta);
  \end{scope}

  \node (sample) [blockStyle,fill=gray!20,above=of delta,yshift=11mm,fill=yellow!20] {Sample HPs};
  \node (spec) [blockStyle,fill=gray!20, left=of sample,fill=red!20] {Specify HP ranges};
  \node (run) [blockStyle,fill=gray!20,right=of sample,fill=green!20] {Run environment with HPs};
  \node (collect) [blockStyle,fill=gray!20,right=of run,fill=blue!20] {Collect EL corpus};
  \draw [-stealth] (spec) to (sample);
  \draw [-stealth] (sample) to (run);
  \draw [-stealth] (run) to (collect);
  \draw [-stealth] (collect) to (beta);
  \draw [-stealth] (delta) to (sample);
  \node (update) at ($(delta)!.65!(sample)$) [
    font=\small,
    fill=white,
    text width=2.5cm,
    draw=black,
    text centered,
  ] {Update sampler};

  \begin{pgfonlayer}{background}
    \node [fill=black!10,fit=(bb),rounded corners,inner sep=1mm] {};
  \end{pgfonlayer}

\end{tikzpicture}
  \caption{Illustration of hyperparameter optimization with XferBench (adapted from \citet{boldt-mortensen-2024-xferbench} (CC BY 4.0 License)).}%
  \label{fig:xb}
\end{figure*}


\section{Methods}%
\label{sec:methods}
\subsection{Objective: XferBench}

The ultimate objective that we are optimizing for is transfer learning performance on downstream human language tasks.
This objective is quantified by XferBench \citep[MIT license]{boldt-mortensen-2024-xferbench}, which measures how much pretraining on an emergent language corpus decreases cross-entropy on a limited-data, downstream language modelling task on human languages (illustrated in the gray box of \cref{fig:xb}).
While language modeling performance does not capture every aspect of mastery of language, it does serve as the backbone of many NLP tasks (e.g., generative models, automatic speech recognition, machine translation).
From a practical point of view, language modeling is also one of the simpler and less expensive downstream tasks to test on (cf.\@ testing on machine translation in \citet{boldt-mortensen-2024-xferbench}).

Since the output of XferBench is mean cross-entropy across human languages, a lower score better.
XferBench takes as input a corpus of $15$ million tokens, which is used for the pretraining stage and finetunes on $2$ million tokens for each evaluation (human) language.
The language model used for XferBench is based on GPT-2 \citep{radford2019language} and has ${\sim}60$ million parameters.
Since XferBench has a long runtime, we use a modified version only during hyperparameter search termed \emph{XferBench-da} which only evaluates on one human language (viz.\@ Danish) which we found to have high correlation ($R^2>0.95$) with the complete XferBench; see \cref{sec:eval-corr} for details.

\subsection{Environment: signalling game}
The environment we use in our experiments is the signalling game.
In particular we use the discrimination variant of the signalling game based on the implementation in EGG \citep[\url{https://github.com/facebookresearch/EGG}, MIT license]{egg}.
The discrimination variant of the signalling game consists of two agents, a sender and a receiver interacting for a single round.
In a given round, the sender observes an input, sends a message  to the receiver, and the receiver selects an observation out of a number of candidates based on the message.
Of the candidate observations, one is correct (i.e., the same as the sender's input), and the rest are ``distractors''.
In the implementation used in this paper:
\begin{itemize}[nosep]
  \item Observations are concatenations of a fixed number of discrete-valued vectors (see \cref{app:obs-rep} for details).
  \item Messages are sequences of integers represented by one-hot vectors.
  \item Agents are feed-forward neural networks with one hidden layer and GRU-based RNNs to generate/read the message.
    \unskip\footnote{Other architectures were investigated in a follow-up experiment described in \cref{sec:arch-scatter}.}
  \item The sender--receiver system is trained end-to-end with backpropagation using a Gumbel-Softmax layer \citep{maddison2017the,jang2017categorical} to generate the message.
\end{itemize}

Overall, this emergent communication system is about as ``vanilla''  as is studied in the literature.
This is advantageous for a number of reasons:
\begin{itemize}[nosep]
  \item The environment is fast to run, requiring $10$ to $120$ minutes depending on the hyperparameters.
  \item It has a (comparatively) limited number of hyperparameters making hyperparameter search more tractable and reducing potential confounding variables.
  \item It serves as a ``lower bound'' for optimizing emergent communication environments since we can determine the maximum performance possible in a system with minimal complexity.
  \item The training is stable, converging to a high success rate for most hyperparameter combinations.
\end{itemize}

The data is generated for the input corpus to XferBench by sampling from the dataset of observations and feeding these observations into the sender which generates the message.

\subsection{Variables: hyperparameters}
The hyperparameters are the independent variable of the primary experiments presented in this paper;
  that is, the hyperparameters will be varied in order to optimize the system for the objective function.
Some hyperparameters manipulated in this study are unique to the signalling game (e.g., how many attributes and values in the signalling game observations) while others come from deep learning-based architectures more generally (e.g., learning rate, neural network architecture).

We primarily investigate the following hyperparameters:
\begin{description}[nosep]
  \item[Learning rate] Multiplication factor for the weight updates for parameters in the neural network.
  \item[Embedding size] Size of embedding layer in both the sender and the receiver networks; these are independent layers, but their sizes are varied in unison for hyperparameter search.
  \item[Hidden size] The size of hidden layer in both the sender and the receiver networks; values are varied in unison.
  \item[\textit{n} attributes] Number of one-hot vectors in each observation.
  \item[\textit{n} values] Size of one-hot vectors in observations.
  \item[\textit{n} distractors] Number of incorrect observations shown to the receiver (in addition to the correct one).
  \item[\textit{n} epochs] Number of training examples seen.
    \unskip\footnote{Since the data is procedurally generated, a new dataset of $1024$ observations is sampled for each epoch.}
  \item[Temperature] Temperature of the Gumbel-Softmax layer which the sender uses to generate messages during training.
  \item[Vocabulary size] Dimension of the one hot vectors which comprise the message.
  \item[Message length] Number of one-hot vectors in a message.\footnote{Technically, the implementation allows for variable length messages, but optimization led to all messages always being the max length.}
\end{description}
Other hyperparameters that were either not discussed or not investigated are documented in \cref{sec:not-discussed}.
Although this set of hyperparameters only covers a small portion of the possible variations of the signalling game (let alone other emergent language games), it covers many basic hyperparameters which show up commonly in emergent communication research.

\begin{table*}
  \newcommand\dit{---}
  \centering
  \small
  \setlength\tabcolsep{0.4em}
  \begin{tabular}{lrrrrrrrrrrr}
    \toprule
    \# & $|\text{Trials}|$ & $|\text{Attrs.}|$ & $|\text{Vals.}|$ & $|\text{Distrs.}|$ & Temp. & $|\text{Embed.}|$ & $|\text{Hidden}|$ & LR & $|\text{Vocab}|$ & Length & $|\text{Epochs}|$ \\
    \midrule
    1  & $578$ & $[3,7]$  &  $[3,7]$ & $[1,127]$ & $[0.1, 10]$ & $[8,128]$  & $[8,128]$  & $[500\text{\textmu},50\text{m}]$ & $[10,20\text{k}]$  & $[1,40]$ & $500$             \\
    2  & $171$ & $[5,10]$ & $[5,10]$ & \dit{}    &  $[0.5, 4]$ & $[64,512]$ & $[64,512]$ & $[500\text{\textmu},5\text{m}]$  & $[300,30\text{k}]$ & \dit{}   & \dit{}            \\
    3  & $140$ & \dit{}   & \dit{}   & \dit{}    & \dit{}      & \dit{}     & \dit{}     & \dit{}                           & \dit{}             & \dit{}   & $[500,5\text{k}]$ \\
    4  & $282$ & $[6,20]$ &    $6$   & $23$      & $2$         & $128$      & $256$      & $[1\text{m},3\text{m}]$          & $[500,30\text{k}]$ & \dit{}   & \dit{}            \\
    4* &   $1$ & $11$     &    $6$   & $23$      & $2$         & $128$      & $256$      & $1.79\text{m}$                           & $9721$            & $16$     & $1715$            \\
    \bottomrule
  \end{tabular}
  \caption{All hyperparameters were treated as log-scale hyperparameters. $|{\cdot}|$ refers to cardinality. ``\dit{}'' means unchanged from the previous run. \textmu, m, and k refer to the SI prefixes micro ($\times10^{-6}$), milli ($\times10^{-3}$), and kilo ($\times10^{3}$), respectively.}%
  \label{tab:hp-search}
\end{table*}

\begin{figure*}
  \centering
  \input{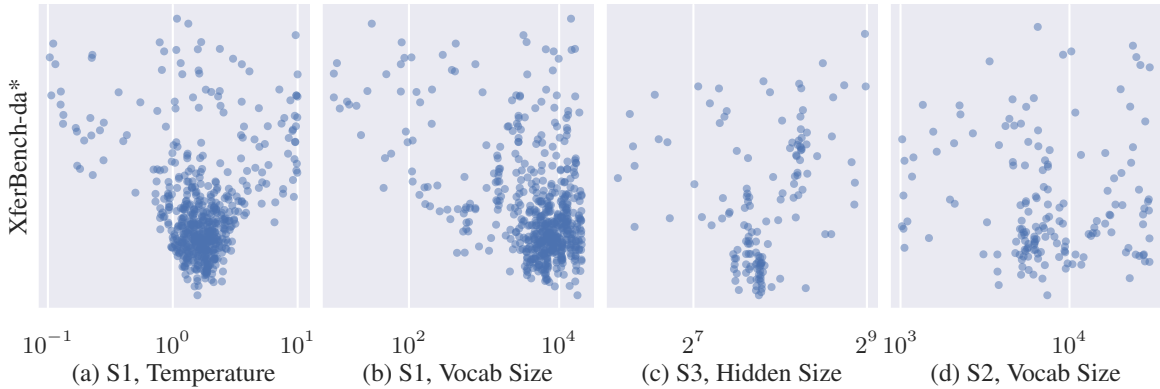}
  \caption{Examples of different hyperparameter--objective relations observed in the various searches and hyperparameters.
    From left-to-right, we have:
      (a) a clear best value,
      (b) a clear trend outside the provided range,
      (c) a weak trend toward a particular value,
      and (d) no definite trend.
    The $y$-axis based on different ``sizes'' of XferBench-da normalized to similar scales.}%
  \label{fig:hpo-slice}
\end{figure*}


\subsection{Optimization: hyperparameter search}
Finally, we discuss the method used for optimizing the hyperparameters of the emergent communication system (the parameters system itself are optimized with backpropagation, as mentioned above).
The simplest of all hyperparameter search methods is grid search, where each element of the Cartesian product of every set of hyperparameter values is evaluated.
Even using a modest $3$ values per aforementioned hyperparameter would require $3^{10}\approx60\,000$ trials, taking $5$ GPU-years (at $1$ hour per trial).
Thus, we employ Bayesian parameter optimization to more efficiently select hyperparameter combinations to evaluate; this additionally allows us to specify a range of hyperparameter values instead of individual values.
This process is illustrated in \cref{fig:xb}.

We specifically use a Tree-structured Parzen Estimator (TPE) \citep{bergstra2011tpe} as implemented in Optuna \citep[MIT license]{optuna}.
At a basic level, TPE works by partitioning hyperparameter combinations into a ``good'' set and a ``bad'' set based on the objective function value and selects the next combination of hyperparameters by maximizing the probability of the hyperparameters being in the good set divided by the probability of them being in the bad set.
These probability estimates use multivariate kernel density estimators and permit discrete, categorical, and conditional hyperparameter values.
After running the environment with the hyperparameters and the objective function on the result, the sampler's probability estimates are updated in accordance with the objective function's value.
For a more detailed explanation, see \citet{watanabe2023tpe-tutorial}.


\section{Experiments}%
\label{sec:experiments}
The code to run the experiments and analyses is publicly available at \url{https://github.com/brendon-boldt/signalling-game-search} under the MIT license.

\subsection{Hyperparameter searches}
In this paper, we present four main searches (Searches 1--4) with two additional searches (Searches 5r and 6e) for use in later analyses (\cref{sec:analysis}).
The following is a summary of the hyperparameter searches:
\smallskip
\begin{description}[nosep,leftmargin=0.2in]
  \item[Search 1] Large number of hyperparameters varied with a wide range; used small version of XferBench-da ($1$M train tokens for $1$ epoch, $200$k test tokens for $2$ epochs).
  \item[Search 2] Same number of hyperparameters varied with smaller or larger ranges depending on results of Search 1; used medium version of XferBench-da ($4$M train tokens for $2$ epochs, $1$M test tokens for $3$ epochs)
  \item[Search 3] Same parameters as Search 2 while allowing number of epochs to go higher and using the full version of XferBench-da ($15$M train tokens for $5$ epochs, $2$M test tokens for $10$ epochs).
  \item[Search 4] Reduces ranges or fixes parameters from Search 3 to maximize exploitation of good parameters; 4* in \cref{tab:hp-search} is the best-performing trial from Search 4.
  \item[Search 5r] Most parameters varied with wide ranges except using \emph{random sampling} to remove sampling bias; similar to Search 1 with narrower ranges on learning rate. Discussed in \cref{sec:ent-xb}.
  \item[Search 6e] Optimized for maximizing entropy after a number of previous searches (not discussed in the paper); similar to Search 4 in this regard. Discussed in \cref{sec:ent-xb}.
\end{description}
The parameters of Searches 1--4 are given in \cref{tab:hp-search} (for complete table, see \cref{tab:hp-search-all}).
The implementation defaults for other hyperparameters were used unless otherwise specified.
Optuna's default parameters for TPE were used across all experiments.

The signalling game takes $5$ to $40$ minutes to run (depending primarily on the number of epochs, and, to a lesser extent, the message length), and the full version of XferBench-da takes approximately $40$ minutes to run.
Thus, the average trial (for the latter searches) takes approximately $[0.75,1.5]$ hours.
Parallelization was used to run multiple trials within a search at a time.
See \cref{sec:resources} for a discussion of computing resources used.

\paragraph{Search design}
For each iteration of the primary searches (i.e., 1--4), we changed the search parameters based on their correlation with the objective function.
We observed four main univariate patterns\footnotemark{}, illustrated in \cref{fig:hpo-slice}.
\footnotetext{While we did look for multivariate effects (i.e., hyperparameters that are \emph{not} independent), we did not observe any notable trends.}
For parameters with a clear trend toward the center (\cref{fig:hpo-slice}a), we narrowed the range to encourage exploiting good values.
Some parameters trended to one side of the range (\cref{fig:hpo-slice}b), which indicated needing to extend the range.
Parameters with weak to no trend (Figures~\ref{fig:hpo-slice}c and~\ref{fig:hpo-slice}d) were left unchanged for the initial searches and given an arbitrary value for the final search to reduce noise.
Full hyperparameter plots given in \cref{sec:hp-scatter}.

Searches 1 and 2 used a reduced version of XferBench to execute more trials quickly and prune the less promising hyperparameter ranges; nevertheless, caution was exercised in pruning since scaling up XferBench could change optimal hyperparameter values.
The irregular number of trials per search were due to executing as many trials as possible within a certain time (rather than aiming for a particular number of trials).

\subsection{Languages evaluated}

We select three categories of languages to evaluate with XferBench:
  human languages, those generated with the hyperparameter search discussed above, and extant emergent language corpora from ELCC \citep[\smallish\url{https://huggingface.co/datasets/bboldt/elcc}, CC BY 4.0]{elcc}.
The primary goal is for the search-derived languages to outperform all existing emergent languages and get as close to human language performance as possible.
For the human languages, we use a subset of the baselines provided in \citet{boldt-mortensen-2024-xferbench}.
In particular, we use Mandarin and Hindi because they were the best- and worst-performing human languages, respectively, and French and Arabic to round out the language families represented.

For the search-derived languages, we selected the three best languages from the final primary run of hyperparameter search (Search 4) and evaluate them on the full set of evaluation languages in XferBench.
We additionally include the three highest-entropy languages from the entropy-maximizing search (Search 6e, discussed further in \cref{sec:ent-xb}).

Finally, for the emergent language-based points of comparison, we select three of the best performing languages from ELCC\@.
Most notably, this includes Yao+ (\texttt{\smallish corpus-transfer-\allowbreak yao-et-al/\mbox{coco\_2014}} \citep{yao2022linking}) which performed far better than all other emergent languages on XferBench.
Mu+ (\texttt{\smallish generalizations-\allowbreak mu-goodman/\allowbreak cub-reference} \citep{mu2021emergent}) and Chaabouni+ (\texttt{\smallish ec-at-scale/imagenet-10x10} \citep{chaabouni2022emergent}) were also included as more typical high-performing emergent languages on XferBench.

\subsection{Results}
\begin{figure}
  \centering
  \input{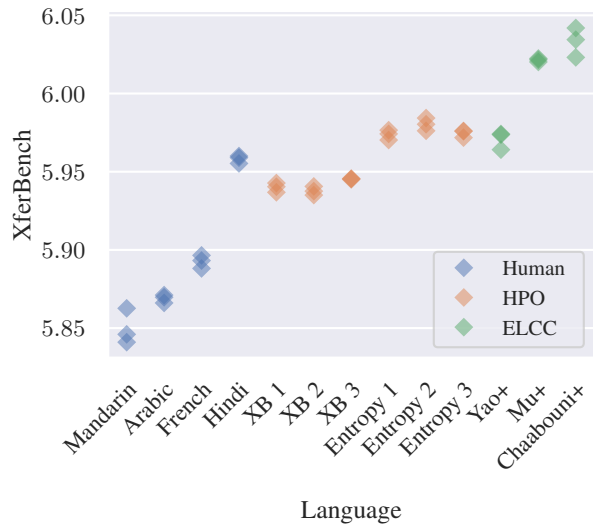}
  \caption{Plot of XferBench scores on emergent and human languages.  XB 1--3 are emergent language corpora derived from Search 4 and Entropy 1--3 from Search 6e.}%
  \label{fig:bar}
\end{figure}

\Cref{fig:bar} shows $3$ randomly seeded runs of the full XferBench score for each corpus.
For the emergent languages from hyperparameter search, the models were restored from checkpoints saved during the search, but the corpora were generated independently of the search.
First, we see that the emergent languages from the XferBench-based search (XB 1--3) outperform all other emergent languages and even the Hindi corpus\footnote{For a brief discussion of Hindi's poor performance, see \cref{sec:hindi-outlier}.}.
While it is indeed significant that these emergent languages outperform a human language corpus, this corpus is also an outlier, and the emergent languages are still relatively far from matching the performance of the rest of the human language corpora.
Nevertheless, these figures show that the XB 1--3 languages achieve state-of-the-art levels of similarity to human language.
The corpora from the entropy-based search (Entropy 1--3) perform well, comparably to Yao+, but significantly worse than the XferBench-search languages.



\section{Analysis}%
\label{sec:analysis}

\subsection{Importance of hyperparameters}

\paragraph{Vocabulary size}
The most notable hyperparameter trend we found was with vocabulary size, where the best-performing languages 
had unique token counts of on the order of $1000$ and vocabulary sizes closer to $10\,000$ (see \cref{fig:slice-3}); that is, the model could use up to $10\,000$ unique words but only uses $1000$ after training.
For reference, it is common practice in emergent communication research to use vocabulary sizes well under $100$ (e.g., only $1$ out of the $8$ systems in ELCC produce corpora with ${>}70$ unique tokens).

\paragraph{Scaling up}
Similarly to vocabulary size, we observe indications to scale up message length, neural network layer size, and task information (i.e., number of attributes, values, and distractors):
  the most human like emergent languages require longer training, larger networks, and higher-information tasks compared to common practice in the emergent communication literature.
Along with vocabulary size, these hyperparameter are most often trivial to adjust, meaning there is little reason not to adjust standard practice in emergent communication research to using hyperparameters in these ranges.

\paragraph{Learning rate}
Finally, in terms of raw importance with respect to XferBench score, learning rate was most significant; this result is not surprising as learning rate is significant in any deep learning algorithm.
Nevertheless, part of the difficulty with learning rate is that there is no one best learning rate, and so performing at least some hyperparameter tuning with learning rate will be necessary for optimal performance.

\paragraph{Summary of recommendations}
We recommend the following hyperparameters as a rule of thumb:
vocabulary size: $10\,000$,
hidden layer size: $256$,
embedding layer size: $128$,
message length: $20$,
observation diversity: the higher the better (e.g., $6^{12}\approx 2\,\text{trillion}$ unique observations),
epochs: train until task success plateau (not just until arbitrary threshold),
learning rate: tune on final setting,
neural architecture: $2$-layer LSTM with $2$ hidden layers\footnote{Based on follow-up experiments in \cref{sec:arch-scatter}.}.

\subsection{Entropy and XferBench}%
\label{sec:ent-xb}

The most striking correlation we observe in our experiments is between XferBench score and unigram token entropy, which is illustrated in \cref{fig:ent-xb} (Pearson's $r=-0.57$ for Search 5r only).
The emergent languages pictured are all those generated by Searches 4 and 5r, while the human languages are taken from \citet{boldt-mortensen-2024-xferbench}. 
We see that low entropy languages tend to score poorly on XferBench while high scoring languages have higher entropy; this aligns with the observed correlation between XferBench and entropy in \citet{elcc}.
Furthermore, this correlation follows the same trend we see in human languages with respect to entropy.

\paragraph{Entropy's lower bound}
In particular, we have illustrated a lower bound of low entropy--low XferBench score that describes both emergent and human languages (the gray dashed line in \cref{fig:ent-xb}).
This suggests that given a certain entropy, there is a hard limit on the performance XferBench that can be achieved.
While further theoretical and empirical analysis would be required to verify that this a true lower bound, this aligns with the notion of language models as entropy-minimizers:
Language models, in order to reduce the entropy on a target language, require a certain degree of entropy (i.e., information) in the pretraining data.
Hence, low-entropy, low-information pretraining data leads to language models which reduce entropy less (i.e., yielding higher cross-entropy).

\paragraph{Entropy minimization}
Looking again at \cref{fig:ent-xb}, we also see that the high-entropy, high-XferBench quadrant (upper right) is also sparsely inhabited.
In fact, emergent and human languages seem to lie primarily near the Pareto frontier of low-entropy, low-XferBench score mentioned above.
This comes in contrast to the XferBench scores of a variety of synthetic languages (descriptions of which are given in \cref{sec:synth}) which often do not demonstrate this Pareto efficiency, even for synthetic languages performing well on XferBench.

This result is concordant with the related claim that entropy is ``minimized'' inside of emergent communication systems \citep{kharitonov2020entmin,chaabouni2021color}.
Such work has shown that emergent communication systems tend to find Pareto efficient solutions in terms of maximizing task success and minimizing entropy (this correlation in the hyperparameter search is discussed briefly in \cref{sec:ent-vs-acc}).

\paragraph{Optimizing on entropy directly}
The correlation between entropy and XferBench naturally leads to a potential performance improvement: Why not use entropy as the hyperparameter objective instead of XferBench?
Entropy takes seconds to compute instead of close to an hour.
This is the experiment performed in Search 6e which was successful in producing languages with good XferBench scores but which still performed significantly worse than optimizing on XferBench directly (see \cref{fig:bar}).

Given that the lower bound of entropy versus XferBench score is tighter than the upper bound, it is roughly the case that low entropy implies poor XferBench performance, but high entropy does not necessarily imply good XferBench performance.
Furthermore, it is also possible that optimizing directly for entropy results in degenerate solutions that find trivial or otherwise unhelpful ways to boost entropy.
Thus, the fact that the entropy-based search finds good but not optimal emergent languages fits with the earlier observation about bounds of entropy and XferBench score.
With these observations in mind, a refinement to the hyperparameter search algorithm would be to prune low-entropy trials before running XferBench while fully evaluating the trial on XferBench if it has high entropy.

\paragraph{Task success}
\begin{figure}
  \centering
  \input{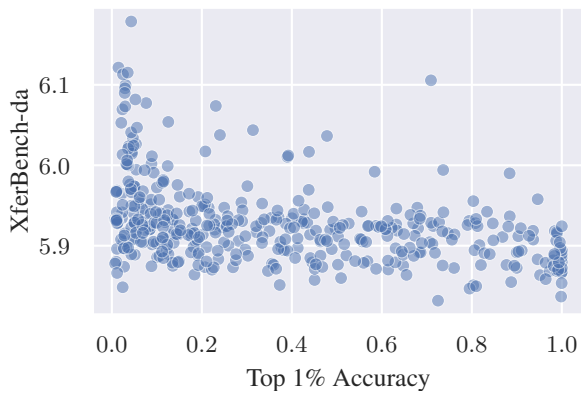}
  \caption{%
    Accuracy versus XferBench for Search 5r.
    Accuracy is measured as proportion of rounds for which the correct observation is ranked in the top-$1$ percentile among all distractors.}%
  \label{fig:acc-vs-xb}
\end{figure}
The correlation between task success and XferBench score (\cref{fig:acc-vs-xb}, Pearson's $r=-0.40$) is not as dramatic as with entropy.
Nevertheless, the negative correlation (better task success, better XferBench score) matches the expectation that the realism of emergent language is positively correlated with the efficacy of the language.
This relationship is a foundational assumption of emergent communication techniques generally: the realism of simulation-derived language comes, in part, from its development out of the functional pressures to communicate.
Thus, if the emergent communication does not function well, we would not have reason to think it would be similar to human language, absent evidence.





\section{Discussion}%
\label{sec:discussion}

\paragraph{Similarity to human language}
The primary motivation for optimizing emergent communication systems on XferBench is to create more human language-like emergent languages.
In this way, this environment and the recommended hyperparameters provide a better baseline environment for future emergent communication research to work from.
This similarity to human language is critical for nearly every application of emergent communication research, not only related to machine learning and NLP but also areas with more linguistic focus \citep{boldt2024review}.
Although XferBench quantifies a decidedly more deep learning, data-driven notion of similarity, this account is complimentary with more explicitly linguistic notions of similarity to human language.

For example, linguistic phenomena such as parts of speech fundamentally concern whole classes of words behaving predictably in a variety of environments.
Thus, trivially small languages are not suitable for addressing such phenomena as there are not classes of words and no variety to generalize over.
Even something as fundamental as the Zipfian distribution of words in human language presupposes a large vocabulary size \citep{zipf1949least,piantadosi2014zipf}.
\unskip\footnote{A follow-up experiment in \cref{sec:zipf} shows that even high-performing emergent languages from our experiments have a decidedly non-Zipfian distribution.}
Furthermore, smaller-scale emergent languages are a greater risk for overfitting since the capacity of a neural network quickly enters the overparameterization regime when the language has as small vocabulary, message length, etc. \citep{gupta-etal-2020-compositionality}.

\paragraph{Emergent properties}
The relationship between entropy, task success, and XferBench score demonstrated in the hyperparameter searches emphasizes the presence of \emph{truly emergent} properties and processes in emergent communication:
Neither entropy nor transfer learning performance are directly optimized for (cf.\@ task success).
Just as Pareto efficient entropy has been found for task success in emergent languages \citep{kharitonov2020entmin}, we find some degree of Pareto efficiency with entropy and XferBench performance (and to a limited degree with task success and XferBench).
What this shows is that the communicative pressures and information theoretic considerations are a key ingredient in emergent language's similarity to human language.
Thus, task success and entropy serve as additional ways to reason about emergent language and how to apply it to human language.
Nevertheless, the limited correlation we find among these properties also tells us that emergent language is not trivially explained by these factors either.

\paragraph{Future work}
On the front of creating more human language-like emergent languages, a next step is to introduce new variations of the signalling game, entirely new environments, or more sophisticated neural architectures and optimize them on a metric like XferBench in order to progress towards the long-term goal of producing realistic emergent languages for transfer learning.
Because this paper has wrung as much performance as is possible from the basic signalling game environment, there can be greater certainty that innovations producing higher-performing languages are actually causing the improvement.
Otherwise, more trivial factors like better learning rate tuning could become confounding variables.

As far as investigating the entropy minimization pressure in emergent languages, further theoretical work needs to build models and generate testable hypotheses; theoretical models are the key to scientific explanation beyond merely showing the existence of correlations.
Nevertheless, this paper has shown that hyperparameter turning can be an effective tool for producing a large variety of emergent language that preclude hyperparameters being confounding variables.
Such methods of generating datasets will be invaluable in empirically testing theoretical models of emergent language.

\section{Conclusion}%
\label{sec:conclusion}
In this paper we have used hyperparameter search to generate the most human language-like emergent language to date, as quantified by XferBench.
Not only does this represent a step forward for using emergent languages as realistic synthetic data for transfer learning but also provides insight into how hyperparameters can be better addressed in future emergent communication research.
Finally, the hyperparameter search reveals further importance of the role of entropy in emergent language.
High entropy appears to be a necessary condition for good transfer learning performance while at the same time, emergent language appears to minimize entropy for a given level of transfer learning performance.
Furthermore, this entropy minimization is not replicated in synthetic languages suggesting that emergent language is more than just ``synthetic languages with extra steps''.

\section*{Limitations}
In terms of finding the most human language-like emergent language, this study is limited in terms of the simplicity of the environment and agent design.
A single round signalling game with a fixed sender and receiver and uniform, synthetic observations is a no-frills environment which, while good for stability and simplicity, is limited in the richness of information to be communicated, and as a result, the languages it can produce.
Thus, while the presented insights can apply, in part, to many settings, it does not come close to providing a comprehensive account of the effects of hyperparameters in emergent communication.

Regarding the investigation of the link between entropy and XferBench score and task success, we were not able to build any theoretical models to scientifically test particular hypotheses about the relationships between the variables; instead, we are only able to offer empirical evidence that there are trends warranting further investigation.
Finally, the recommendations we can given regarding the hyperparameters of emergent communication systems are limited because hyperparameter search is relatively ``messy''; it is geared toward maximizing performance more than uncovering generalizable trends.
Additionally, we perform our experiments with a signalling game which provides only limited evidence for the behavior of emergent communication systems with different tasks.


\bibliography{src/main}

\begin{thebibliography}{26}
\providecommand{\natexlab}[1]{#1}

\bibitem[{Akiba et~al.(2019)Akiba, Sano, Yanase, Ohta, and Koyama}]{optuna}
Takuya Akiba, Shotaro Sano, Toshihiko Yanase, Takeru Ohta, and Masanori Koyama. 2019.
\newblock {O}ptuna: A next-generation hyperparameter optimization framework.
\newblock In \emph{The 25th ACM SIGKDD International Conference on Knowledge Discovery \& Data Mining}, pages 2623--2631.

\bibitem[{Bergstra et~al.(2011)Bergstra, Bardenet, Bengio, and K\'{e}gl}]{bergstra2011tpe}
James Bergstra, R\'{e}mi Bardenet, Yoshua Bengio, and Bal\'{a}zs K\'{e}gl. 2011.
\newblock \href {https://proceedings.neurips.cc/paper_files/paper/2011/file/86e8f7ab32cfd12577bc2619bc635690-Paper.pdf} {Algorithms for hyper-parameter optimization}.
\newblock In \emph{Advances in Neural Information Processing Systems}, volume~24. Curran Associates, Inc.

\bibitem[{Boldt and Mortensen(2023)}]{boldt2023mathmodel}
Brendon Boldt and David Mortensen. 2023.
\newblock \href {https://arxiv.org/abs/2211.15783} {Mathematically modeling the lexicon entropy of emergent language}.
\newblock \emph{arXiv}, 2211.15783.

\bibitem[{Boldt and Mortensen(2024{\natexlab{a}})}]{elcc}
Brendon Boldt and David Mortensen. 2024{\natexlab{a}}.
\newblock \href {https://arxiv.org/abs/2407.04158} {{ELCC}: the {E}mergent {L}anguage {C}orpus {C}ollection}.
\newblock \emph{Preprint}, arXiv:2407.04158.

\bibitem[{Boldt and Mortensen(2024{\natexlab{b}})}]{boldt-mortensen-2024-xferbench}
Brendon Boldt and David Mortensen. 2024{\natexlab{b}}.
\newblock \href {https://doi.org/10.18653/v1/2024.naacl-long.82} {{X}fer{B}ench: a data-driven benchmark for emergent language}.
\newblock In \emph{Proceedings of the 2024 Conference of the North American Chapter of the Association for Computational Linguistics: Human Language Technologies (Volume 1: Long Papers)}, pages 1475--1489, Mexico City, Mexico. Association for Computational Linguistics.

\bibitem[{Boldt and Mortensen(2024{\natexlab{c}})}]{boldt2024review}
Brendon Boldt and David~R Mortensen. 2024{\natexlab{c}}.
\newblock \href {https://openreview.net/forum?id=jesKcQxQ7j} {A review of the applications of deep learning-based emergent communication}.
\newblock \emph{Transactions on Machine Learning Research}.

\bibitem[{Chaabouni et~al.(2021)Chaabouni, Kharitonov, Dupoux, and Baroni}]{chaabouni2021color}
Rahma Chaabouni, Eugene Kharitonov, Emmanuel Dupoux, and Marco Baroni. 2021.
\newblock \href {https://doi.org/10.1073/pnas.2016569118} {Communicating artificial neural networks develop efficient color-naming systems}.
\newblock \emph{Proceedings of the National Academy of Sciences}, 118(12):e2016569118.

\bibitem[{Chaabouni et~al.(2022)Chaabouni, Strub, Altch{\'e}, Tarassov, Tallec, Davoodi, Mathewson, Tieleman, Lazaridou, and Piot}]{chaabouni2022emergent}
Rahma Chaabouni, Florian Strub, Florent Altch{\'e}, Eugene Tarassov, Corentin Tallec, Elnaz Davoodi, Kory~Wallace Mathewson, Olivier Tieleman, Angeliki Lazaridou, and Bilal Piot. 2022.
\newblock \href {https://openreview.net/forum?id=AUGBfDIV9rL} {Emergent communication at scale}.
\newblock In \emph{International Conference on Learning Representations}.

\bibitem[{Cho et~al.(2014)Cho, van Merri{\"e}nboer, Bahdanau, and Bengio}]{gru}
Kyunghyun Cho, Bart van Merri{\"e}nboer, Dzmitry Bahdanau, and Yoshua Bengio. 2014.
\newblock \href {https://doi.org/10.3115/v1/W14-4012} {On the properties of neural machine translation: Encoder{--}decoder approaches}.
\newblock In \emph{Proceedings of {SSST}-8, Eighth Workshop on Syntax, Semantics and Structure in Statistical Translation}, pages 103--111, Doha, Qatar. Association for Computational Linguistics.

\bibitem[{Elman(1990)}]{elman}
Jeffrey~L. Elman. 1990.
\newblock \href {https://doi.org/10.1207/s15516709cog1402\_1} {Finding structure in time}.
\newblock \emph{Cognitive Science}, 14(2):179--211.

\bibitem[{Gupta et~al.(2020)Gupta, Resnick, Foerster, Dai, and Cho}]{gupta-etal-2020-compositionality}
Abhinav Gupta, Cinjon Resnick, Jakob Foerster, Andrew Dai, and Kyunghyun Cho. 2020.
\newblock \href {https://doi.org/10.18653/v1/2020.repl4nlp-1.5} {Compositionality and capacity in emergent languages}.
\newblock In \emph{Proceedings of the 5th Workshop on Representation Learning for NLP}, pages 34--38, Online. Association for Computational Linguistics.

\bibitem[{Hochreiter and Schmidhuber(1997)}]{lstm}
Sepp Hochreiter and J\"urgen Schmidhuber. 1997.
\newblock \href {https://doi.org/10.1162/neco.1997.9.8.1735} {Long short-term memory}.
\newblock \emph{Neural Computation}, 9(8):1735--1780.

\bibitem[{Jang et~al.(2017)Jang, Gu, and Poole}]{jang2017categorical}
Eric Jang, Shixiang Gu, and Ben Poole. 2017.
\newblock \href {https://openreview.net/forum?id=rkE3y85ee} {Categorical reparameterization with gumbel-softmax}.
\newblock In \emph{International Conference on Learning Representations}.

\bibitem[{Kharitonov et~al.(2020)Kharitonov, Chaabouni, Bouchacourt, and Baroni}]{kharitonov2020entmin}
Eugene Kharitonov, Rahma Chaabouni, Diane Bouchacourt, and Marco Baroni. 2020.
\newblock \href {https://proceedings.mlr.press/v119/kharitonov20a.html} {Entropy minimization in emergent languages}.
\newblock In \emph{Proceedings of the 37th International Conference on Machine Learning}, volume 119 of \emph{Proceedings of Machine Learning Research}, pages 5220--5230. PMLR.

\bibitem[{Kharitonov et~al.(2021)Kharitonov, Dess{\`i}, Chaabouni, Bouchacourt, and Baroni}]{egg}
Eugene Kharitonov, Roberto Dess{\`i}, Rahma Chaabouni, Diane Bouchacourt, and Marco Baroni. 2021.
\newblock {EGG}: a toolkit for research on {E}mergence of lan{G}uage in {G}ames.
\newblock \url{https://github.com/facebookresearch/EGG}.

\bibitem[{Lazaridou and Baroni(2020)}]{lazaridou2020emergentmultiagentcommunicationdeep}
Angeliki Lazaridou and Marco Baroni. 2020.
\newblock \href {https://arxiv.org/abs/2006.02419} {Emergent multi-agent communication in the deep learning era}.
\newblock \emph{Preprint}, arXiv:2006.02419.

\bibitem[{Maddison et~al.(2017)Maddison, Mnih, and Teh}]{maddison2017the}
Chris~J. Maddison, Andriy Mnih, and Yee~Whye Teh. 2017.
\newblock \href {https://openreview.net/forum?id=S1jE5L5gl} {The concrete distribution: A continuous relaxation of discrete random variables}.
\newblock In \emph{International Conference on Learning Representations}.

\bibitem[{Mu and Goodman(2021)}]{mu2021emergent}
Jesse Mu and Noah Goodman. 2021.
\newblock \href {https://openreview.net/forum?id=yq5MYHVaClG} {Emergent communication of generalizations}.
\newblock In \emph{Advances in Neural Information Processing Systems}.

\bibitem[{Piantadosi(2014)}]{piantadosi2014zipf}
S.T. Piantadosi. 2014.
\newblock \href {https://doi.org/10.3758/s13423-014-0585-6} {Zipf’s word frequency law in natural language: A critical review and future directions}.
\newblock \emph{Psychon Bull Rev}, 21:1112--–1130.

\bibitem[{Radford et~al.(2019)Radford, Wu, Child, Luan, Amodei, and Sutskever}]{radford2019language}
Alec Radford, Jeff Wu, Rewon Child, David Luan, Dario Amodei, and Ilya Sutskever. 2019.
\newblock Language models are unsupervised multitask learners.

\bibitem[{Scholz et~al.(2024)Scholz, Pelletier, Pullum, and Nefdt}]{sep-linguistics}
Barbara~C. Scholz, Francis~Jeffry Pelletier, Geoffrey~K. Pullum, and Ryan Nefdt. 2024.
\newblock {Philosophy of Linguistics}.
\newblock In Edward~N. Zalta and Uri Nodelman, editors, \emph{The {Stanford} Encyclopedia of Philosophy}, {S}pring 2024 edition. Metaphysics Research Lab, Stanford University.

\bibitem[{Sch\"utzenberger(1963)}]{schutzenberger1963}
M.P. Sch\"utzenberger. 1963.
\newblock \href {https://doi.org/10.1016/S0019-9958(63)90306-1} {On context-free languages and push-down automata}.
\newblock \emph{Information and Control}, 6(3):246--264.

\bibitem[{Suzgun et~al.(2019)Suzgun, Belinkov, Shieber, and Gehrmann}]{suzgun-etal-2019-lstm}
Mirac Suzgun, Yonatan Belinkov, Stuart Shieber, and Sebastian Gehrmann. 2019.
\newblock \href {https://doi.org/10.18653/v1/W19-3905} {{LSTM} networks can perform dynamic counting}.
\newblock In \emph{Proceedings of the Workshop on Deep Learning and Formal Languages: Building Bridges}, pages 44--54, Florence. Association for Computational Linguistics.

\bibitem[{Watanabe(2023)}]{watanabe2023tpe-tutorial}
Shuhei Watanabe. 2023.
\newblock \href {https://arxiv.org/abs/2304.11127} {Tree-structured parzen estimator: Understanding its algorithm components and their roles for better empirical performance}.
\newblock \emph{arXiv}, 2304.11127.

\bibitem[{Yao et~al.(2022)Yao, Yu, Zhang, Narasimhan, Tenenbaum, and Gan}]{yao2022linking}
Shunyu Yao, Mo~Yu, Yang Zhang, Karthik~R Narasimhan, Joshua~B. Tenenbaum, and Chuang Gan. 2022.
\newblock \href {https://openreview.net/forum?id=49A1Y6tRhaq} {Linking emergent and natural languages via corpus transfer}.
\newblock In \emph{International Conference on Learning Representations}.

\bibitem[{Zipf(1949)}]{zipf1949least}
GK~Zipf. 1949.
\newblock \emph{Human behavior and the principle of least effort}.
\newblock Addison-Wesley, Cambridge, MA.

\end{thebibliography}

\appendix
\crefalias{section}{appendix}

\section{Correlation of Evaluation Languages}%
\label{sec:eval-corr}

\begin{table}
  \centering
  \begin{tabular}{lrrr}
\toprule
 & All & Human & Emergent \\
\midrule
Basque & \cellcolor{blue!50!white!3.344426650115238!orange!50!white} $0.340$ & \cellcolor{blue!50!white!54.5041897441573!orange!50!white} $0.685$ & \cellcolor{blue!50!white!0.0!orange!50!white} $0.318$ \\
Danish & \cellcolor{blue!50!white!100.00000000000001!orange!50!white} $0.992$ & \cellcolor{blue!50!white!96.15409362767512!orange!50!white} $0.966$ & \cellcolor{blue!50!white!99.21019572054269!orange!50!white} $0.987$ \\
Finnish & \cellcolor{blue!50!white!96.7998055326322!orange!50!white} $0.971$ & \cellcolor{blue!50!white!96.33650140179637!orange!50!white} $0.968$ & \cellcolor{blue!50!white!96.58622744025485!orange!50!white} $0.969$ \\
Hebrew & \cellcolor{blue!50!white!96.29367056163244!orange!50!white} $0.967$ & \cellcolor{blue!50!white!96.22918064751217!orange!50!white} $0.967$ & \cellcolor{blue!50!white!97.77386416288793!orange!50!white} $0.977$ \\
Indonesian & \cellcolor{blue!50!white!99.41329359994668!orange!50!white} $0.988$ & \cellcolor{blue!50!white!94.06195630718246!orange!50!white} $0.952$ & \cellcolor{blue!50!white!98.5928100591212!orange!50!white} $0.983$ \\
Japanese & \cellcolor{blue!50!white!97.09172096597844!orange!50!white} $0.973$ & \cellcolor{blue!50!white!90.82334928727873!orange!50!white} $0.930$ & \cellcolor{blue!50!white!97.2644629436565!orange!50!white} $0.974$ \\
Kazakh & \cellcolor{blue!50!white!98.6701216187445!orange!50!white} $0.983$ & \cellcolor{blue!50!white!91.66276507427317!orange!50!white} $0.936$ & \cellcolor{blue!50!white!97.70894443495052!orange!50!white} $0.977$ \\
Persian & \cellcolor{blue!50!white!96.97566971433152!orange!50!white} $0.972$ & \cellcolor{blue!50!white!93.94881987525477!orange!50!white} $0.951$ & \cellcolor{blue!50!white!96.91100709006838!orange!50!white} $0.971$ \\
Romanian & \cellcolor{blue!50!white!98.9329710758789!orange!50!white} $0.985$ & \cellcolor{blue!50!white!92.93796160260834!orange!50!white} $0.945$ & \cellcolor{blue!50!white!98.49385441089474!orange!50!white} $0.982$ \\
Urdu & \cellcolor{blue!50!white!93.83569399785327!orange!50!white} $0.951$ & \cellcolor{blue!50!white!78.78049621838137!orange!50!white} $0.849$ & \cellcolor{blue!50!white!90.67773590251264!orange!50!white} $0.929$ \\
\bottomrule
\end{tabular}

  \caption{$R^2$ values for individual target XferBench languages predicting the full XferBench score.  \emph{Human} and \emph{Emergent} refer to the $R^2$ value considering only the human or emergent languages, respectively.}%
  \label{tab:target-corr}
\end{table}
One of XferBench's chief weaknesses is its long runtime, taking $2$ to $6$ hours depending on the GPU used.
Approximately $30\%$ of that time is spent on the initial pretraining with the emergent language corpus, with the other $70\%$ spent on finetuning and testing on the $10$ downstream languages.
We observe from the XferBench scores on the emergent languages of ELCC and the human language baselines of \citet{boldt-mortensen-2024-xferbench} that $9$ out of the $10$ evaluation languages are highly correlated with each other, that is, the XferBench score on one language is highly predictive of the overall XferBench score.
In particular, test cross-entropy on Danish (da) alone can predict ${>}95\%$ of the variation of the overall XferBench score (i.e., the linear regression has an $R^2>0.95$).
For this reason, in the hyperparameter optimization trials, we compute XferBench-da (XferBench evaluated on Danish only) which is around $3{\times}$ faster than the full XferBench; the final evaluation nevertheless uses the full set of evaluation language for XferBench.

In \cref{tab:target-corr}, we show the $R^2$ values derived from training a linear model on just one of the target language's XferBench scores to predict the overall XferBench score.
The emergent languages are all of the corpora from ELCC \citep{elcc}, and the human language corpora are the baselines from the original XferBench paper \citep{boldt-mortensen-2024-xferbench}.
$R^2$ value corresponds to the percent of the variance in the full XferBench score explained by just the score (i.e., cross-entropy) on that particular target language.
We find, strikingly enough, that all of the target languages, with the exception of Basque, are highly correlated, having $R^2$ values above $0.95$ all languages, and greater than $0.80$ even when considering human languages alone.
Danish, of all of the languages, has the highest $R^2$ value (${>}0.99$), which is the reason we select it as the sole target for a more time-efficient variant of XferBench (which we term XferBench-da).

\section{Representation of Signalling Game Observations}%
\label{app:obs-rep}
The originally intended representation for observations in the signalling game was to concatenate one-hot vectors each of which represented the value of of one attribute.
For example, the $2$-attribute, $3$-value vector $[1,2]$, would be represented as $[0,1,0,0,0,1]$ with the first three entries corresponding to the first attribute the last three corresponding to the second attribute.
Due to a mistake in the implementation, the actual representation used was simply a vector of the raw integer values such that $[1,2]$ was simply represented as $[1,2]$.
This to say instead of observations being elements of $\{0,1\}^{|A|\cdot|V|}$ as originally intended, they were implemented as elements of $\mathbb Z^{|A|}$.
The agents did not seem to struggle playing this signalling game even with higher numbers of values.

\section{Hyperparameters Not Discussed}%
\label{sec:not-discussed}
In this section we briefly discuss hyperparameters that were tried but not not documented in the paper or that were not investigated at all.
We selected a batch size of $32$ based on comparing the compute efficiency of different sizes.
Larger batch sizes could process more data faster but would not update the parameters often enough.
On the other hand, smaller batch sizes would not process enough data to maximize the utility of each update.
Mixed precision training was tested but not found to improve runtime.
For learning rate scheduling, we found cosine annealing to be slightly more effective than no schedule, but further schedules were not investigated.
Weight decay was investigated in earlier experiment but found not to have a noticeable effect.

The implementation of the signalling game we used could also be optimized using REINFORCE to handle the discrete message, but we only tested with a Gumbel-Softmax layer as it is faster and more stable to optimize with.

\section{Full Table of Hyperparameters}%
\label{sec:all-hparams}

In \cref{tab:hp-search-all}, we show all of the hyperparameters selected for the searches and trials referenced in the paper.

\begin{table*}
  \newcommand\dit{---}
  \centering
  \small
  \setlength\tabcolsep{0.4em}
  \begin{tabular}{lrrrrrrrrrrr}
    \toprule
    \# & $|\text{Trials}|$ & $|\text{Attrs.}|$ & $|\text{Vals.}|$ & $|\text{Distrs.}|$ & Temp. & $|\text{Embed.}|$ & $|\text{Hidden}|$ & LR & $|\text{Vocab}|$ & Length & $|\text{Epochs}|$ \\
    \midrule
    1    & $578$ & $[3,7]$  &  $[3,7]$ & $[1,127]$  & $[0.1, 10]$ & $[8,128]$  & $[8,128]$  & $[500\text{\textmu},50\text{m}]$ & $[10,20\text{k}]$  & $[1,40]$ & $500$             \\
    2    & $171$ & $[5,10]$ & $[5,10]$ & \dit{}     &  $[0.5, 4]$ & $[64,512]$ & $[64,512]$ & $[500\text{\textmu},5\text{m}]$  & $[300,30\text{k}]$ & \dit{}   & \dit{}            \\
    3    & $140$ & \dit{}   &   \dit{} & \dit{}     & \dit{}      & \dit{}     & \dit{}     & \dit{}                           & \dit{}             & \dit{}   & $[500,5\text{k}]$ \\
    4    & $282$ & $[6,20]$ &      $6$ & $23$       & $2$         & $128$      & $256$      & $[1\text{m},3\text{m}]$          & $[500,30\text{k}]$ & \dit{}   & \dit{}            \\
    4.1  & $1$   & $11$     &      $6$ & \dit{}     & \dit{}      & \dit{}     & \dit{}     & $1.79\text{m}$                   & $9721$             & $16$     & $1715$            \\
    4.2  & $1$   & $12$     &      $6$ & \dit{}     & \dit{}      & \dit{}     & \dit{}     & $1.86\text{m}$                   & $12496$            & $22$     & $1593$            \\
    4.3  & $1$   & $13$     &      $6$ & \dit{}     & \dit{}      & \dit{}     & \dit{}     & $1.74\text{m}$                   & $8096$             & $18$     & $1511$            \\
    5r   & $411$ & $[4,20]$ & $[3,10]$ & $[1,127]$  & $[0.1,10]$  & $[8,512]$  & $[8,512]$  & $[500\text{\textmu},10\text{m}]$ & $[2,30\text{k}]$   & $[1,40]$ & $[10,3\text{k}]$  \\
    6e   & $109$ & $10$     &     $10$ & $[63,511]$ & $2$         & $32$       & $32$       & $2.7\text{m}$                    & $25\text{k}$       & $15$     & $5\text{k}$       \\
    6e.1 & $1$   & \dit{}   &   \dit{} & $228$      & \dit{}      & \dit{}     & \dit{}     & \dit{}                           & \dit{}             & \dit{}   & \dit{}            \\
    6e.2 & $1$   & \dit{}   &   \dit{} & $372$      & \dit{}      & \dit{}     & \dit{}     & \dit{}                           & \dit{}             & \dit{}   & \dit{}            \\
    6e.2 & $1$   & \dit{}   &   \dit{} & $165$      & \dit{}      & \dit{}     & \dit{}     & \dit{}                           & \dit{}             & \dit{}   & \dit{}            \\
    \bottomrule
  \end{tabular}
  \caption{All hyperparameters were treated as log-scale hyperparameters. $|{\cdot}|$ refers to cardinality. ``\dit{}'' means unchanged from the previous run. \textmu, m, and k refer to the SI prefixes micro ($\times10^{-6}$), milli ($\times10^{-3}$), and kilo ($\times10^{3}$), respectively.  4.1 is the best-performing trial of Search 4 (and likewise for 4.2, 6e.1, etc.).}%
  \label{tab:hp-search-all}
\end{table*}

\section{Computing Resources Used}%
\label{sec:resources}
Experiments were performed across about $20$--$30$ NVIDIA A6000 (or equivalent) GPUs (one trial per GPU) on an institutional cluster.
We estimate approximately $5500$ GPU-hours were used for all experiments directly related to this paper, including those not documented or directly referenced.
The primary searches for the best-performing emergent languages on XferBench (Searches 1--4) took about $1300$ GPU-hours.

\section{Hindi's Outlying Score on XferBench}%
\label{sec:hindi-outlier}
Both in this paper as well as the original XferBench paper \citep{boldt-mortensen-2024-xferbench}, Hindi appears to be an outlier in terms of XferBench performance compared to other human languages.
\emph{A priori}, we do not have any reason to expect this, especially since the embeddings (and hence lexical information) are not transferred from training to tuning in XferBench.
\emph{A posteriori}, we can see that based on the entropy of the Hindi corpus from the XferBench baselines, Hindi's poor performance is \emph{not} an outlier as it follows the trend depicted in \cref{fig:ent-xb} (it is the cluster of green square points along the bottom of the cluster of blue circular points); that is, Hindi's entropy of ${\sim}7$ bits is unexpectedly low for human language (cf. ${\sim}11$ bits), but given this low entropy, it performs as expected on XferBench.

\begin{table}
\centering
  \begin{tabular}{lrrr}
  \toprule
  Lang 	& Byte BPETs &	Char BPETs & Char XB\\
  \midrule
  zh 	& $470\,\text{k}$ &  $950\,\text{k}$   & $5.95$ \\
  fr 	  & $900\,\text{k}$ &  $900\,\text{k}$ & $5.95$ \\
  hi 	  &$2100\,\text{k}$ &  $840\,\text{k}$ & $5.94$ \\
  \bottomrule
  \end{tabular}
  \caption{BPE token counts for a parallel corpus in various languages and encoding methods.  XferBench score with character-level BPE training corpus also provided.}%
  \label{tab:char-enc}
\end{table}

While a general data quality program could be causing this low entropy (although Wikipedia data should be relatively clean), we also suspected an encoding problem for Hindi, in part because it is the only baseline language using the particular script (viz. Devanagari, although we do not have a guess why this script would be problematic as compared to others.).
Thus, in an informal follow-up experiment, we encoded a parallel text in Mandarin, French, and Hindi (best-, middle-, and worst-performing languages) using both byte-level BPE as well as character-level BPE.
The results in \cref{tab:char-enc} show that Mandarin is the most efficient for encoding the corpus with byte-level BPE tokens followed by French with Hindi taking more than double the tokens of French.
Since XferBench is token-limited, taking more tokens to represent the same data effectively lowers the amount of data that the language model trains on for Hindi, which has a negative effect on downstream performance (i.e., the XferBench score).
Using character-level BPE instead yields similar corpus sizes, and, indeed, running XferBench with character-level BPE during training yields similar scores for all three languages (although they have all regressed to Hindi's byte-level BPE performance).
Additionally, running XferBench with character-level BPE training led to instabilities with $1$ out of $3$ runs extremely poor performance, possibly due to character-level BPE being more sensitive to the complete set of characters is the training and tuning corpora.

\section{Synthetic Languages}%
\label{sec:synth}
\subsection{Definitions}
We use four probabilistic synthetic languages which span a large portion of the Chomsky hierarchy ranging from trivial to beyond context-free.
All synthetic languages contain a unique begin- and end-of-sentence token in each utterance.

\paragraph{Zipf-Mandelbrot Distribution}
The basis for our synthetic languages will be a Zipf--Mandelbrot distribution, a generalization of Zipf's law, where the unnormalized probability weight of the word $w_i$ is
\begin{equation}
  f(w_i) = \frac1{(i+\beta)^\alpha}
  ,
\end{equation}
where
  $i$ is the $1$-based index of the word,
  $\alpha$ controls the weight of the tail,
  and $\beta$ shifts where the distribution starts (roughly speaking).
Empirically, $\alpha=1$ and $\beta=2.7$ have been found to be good approximations for human language and will be the default parameters of the distribution unless otherwise specified \citep{piantadosi2014zipf}.

\paragraph{Bag of Words}
The simplest synthetic language we introduce is a bag-of-words language where each token in a sentence is sampled independently from the Zipf-Mandelbrot distribution.
The length of the sentence is independent of the sampling method, so in interest of simplicity, we sample from a discrete uniform distribution.

\paragraph{Regular}
The simplest non-trivial language we introduce is a regular language which partitions the tokens uniformly at random into $k$ different sets ($s_1,\dots,s_k$), keeping their initial Zipf--Mandelbrot-derived weight.
Each sentence starts with a token sampled from $s_1$; each subsequent token is sampled from the next class ($s_i+1$) with probability $c$ or sampled from the same class ($s_i$).
After $s_k$, the sentence terminates.
Thus, the language is defined by the regular expression
\begin{equation}
  s_1^+
  s_2^+
  \dots
  s_k^+
  ,
\end{equation}
where
  $a^+=aa^*$,
  $s_i$ represents any token in the set $s_i$,
  and appropriate BoS and EoS tokens are added.

\paragraph{Dyck-$\textit{n}$}
Dyck-$n$ can be thought of as ``balanced nested delimiters'' (where the delimiters are the same token) \citep{schutzenberger1963}.
Each token in the sentence is generated as follows:
  With probability $p$, a new token is sampled from the Zipf--Mandelbrot distribution and pushed onto a stack (the ``opening delimiter''), and with probability $1-p$, the token on top of the stack is popped off.
A sentence always begins with an ``open'' token and ends when the stack is empty.
An example of such a sentence is $(3, 1, 1, 2, 1, 1, 2, 3)$ which could be illustrated as ``\{()[()]\}''.

\paragraph{Shuffle Dyck-$\textit{n}$}
Finally, we use Shuffle Dyck-$n$ as our last language which lies beyond context-free in the Chomsky hierarchy \citet{suzgun-etal-2019-lstm}.
Technically speaking, this language should be called Shuffle of $n$ Distinct Dyck-$1$ Languages since it is the result of randomly interleaving multiple Dyck-$1$ languages with distinct tokens.
To generate a sentence in Shuffle Dyck-$n$, we first follow the same procedure as for Dyck-$n$ but keep the individual tokens separate.
We then interleave the separate strings by appending to the sentence uniformly at random from one of the individual strings until they are empty.
For example, if Dyck-$n$ generated ``\{([()])[]\}'', the separated strings would be ``\{\}'', ``(())'', and ``[][]'', which could then be interleaved into ``\{[\}(()])''.

\subsection{Hyperparameters}

Each variation of the synthetic language maintains the default values while varying a single hyperparameter.
We vary the common hyperparameters as follows:
\begin{description}
  \item[Vocabulary size]
    takes the values $10$, $100$, $1$k, $5$k, $10$k, $30$k (default: $30$k).  A vocab size of $10$ is incompatible with the Regular language and was skipped.
  \item[Zipf--Mandelbrot $\alpha$]
    takes the values $0$, $0.25$, $0.5$, $1$, $2$, and $4$ (default: $1$).
  \item[\textit{n} tokens]
    (in the whole corpus) takes the values $1$k, $10$k, $100$k, $1$M, $5$M, and $15$M (default: $15$M); this hyperparameter was not varied for the Unigram language.
\end{description}

The Unigram language has an additional hyperparameter stop probability which takes the values $0.05$, $0.1$, and $0.2$ (default: $0.1$).
The Regular language has two additional hyperparameters: repeat probability ($c$) which takes the values $0.2$, $0.4$, $0.5$, and $0.6$ (default: $0.4$), and $n$ classes which takes the values $5$, $10$, $20$, and $40$ (default: $10$).
The Dyck and Shuffle Dyck languages take the additional hyperparameter open probability with values: $0.2$, $0.3$, $0.4$, $0.5$, and $0.6$ (default: $0.5$); Shuffle Dyck is not generated with the value $0.6$ due to implementation constraints.

\section{Task Success and Entropy}%
\label{sec:ent-vs-acc}
\begin{figure}
  \centering
  \input{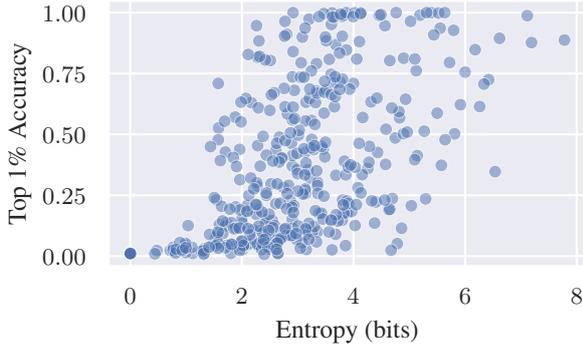}
  \caption{Entropy versus accuracy for Search 5r.}%
  \label{fig:ent-vs-acc}
\end{figure}

Previous work \citep{kharitonov2020entmin,chaabouni2021color} has analyzed entropy minimization with respect to the amount of information or, roughly speaking, task success.
We performed a brief analysis the relationship between entropy and accuracy (task success) shown in \cref{fig:ent-vs-acc}.
While we do find significant correlation (Pearson's $r=0.57$ for Search 5r), we would not characterize it as any strict sort of entropy minimization.
That is, we observe many emergent languages which are from the Pareto frontier of high accuracy and low entropy.
Hyperparameter search demonstrates itself to be a powerful tool for investigating such correlations since it is able to generate a wide variety of emergent languages with minimal additional work from the researchers.
Nevertheless, more investigation would have to be done on this front to conclusively support or reject prior claims of entropy minimization.

\section{Rank--Frequency Plots}%
\label{sec:zipf}

\begin{figure}
  \centering
  \input{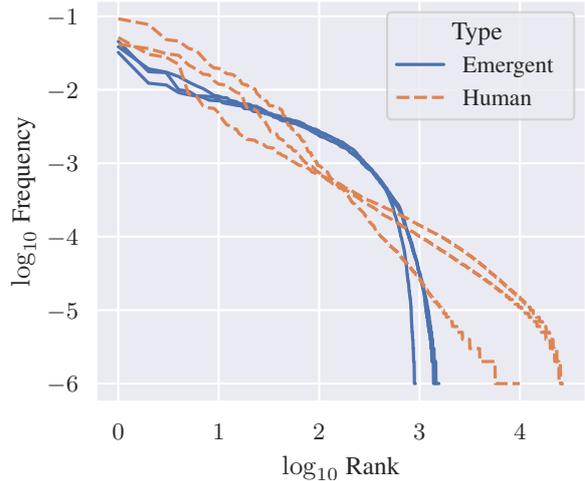}
  \caption{Token log rank versus log frequency plots for emergent and human languages.  Logarithms are in base $10$.}%
  \label{fig:zipf}
\end{figure}

\Cref{fig:zipf} shows Zipf's Law--style plots of rank versus frequency on a log--log scaled plot \citep{zipf1949least} for human languages and high-performing emergent languages.
As Zipf's Law predicts, the human languages show a roughly linear relationship in log--log space.
On the other hand, the emergent languages exhibit more of a ``cliff'' where higher-ranked tokens have a more similar frequency before quickly falling to near-zero frequency.
This implies that human language displays a long tail which is not present in the emergent languages.
The fact that the emergent languages studied exhibit this behavior is somewhat expected as the underlying data distribution that they are representing is itself uniform.


\section{Hyperparameter Scatter Plots}%
\label{sec:hp-scatter}

\Cref{fig:slice-0,fig:slice-1,fig:slice-2,fig:slice-3} show the univariate scatter plots for hyperparameter Searches 1--4.
The $y$-axis is XferBench-da score (or some smaller variant thereof, for Searches 1 and 2), and the $x$-axis is one of the hyperparameters varied for that search.
Note that other variables are \emph{not} held constant while one is varied; instead all hyperparameters are varied for each trial.

\newcommand\slicefig[2]{%
\begin{figure*}
  \centering
  \includegraphics{assets/slices-#1.pdf}
  \caption{#2}%
  \label{fig:slice-#1}
\end{figure*}
}
\slicefig{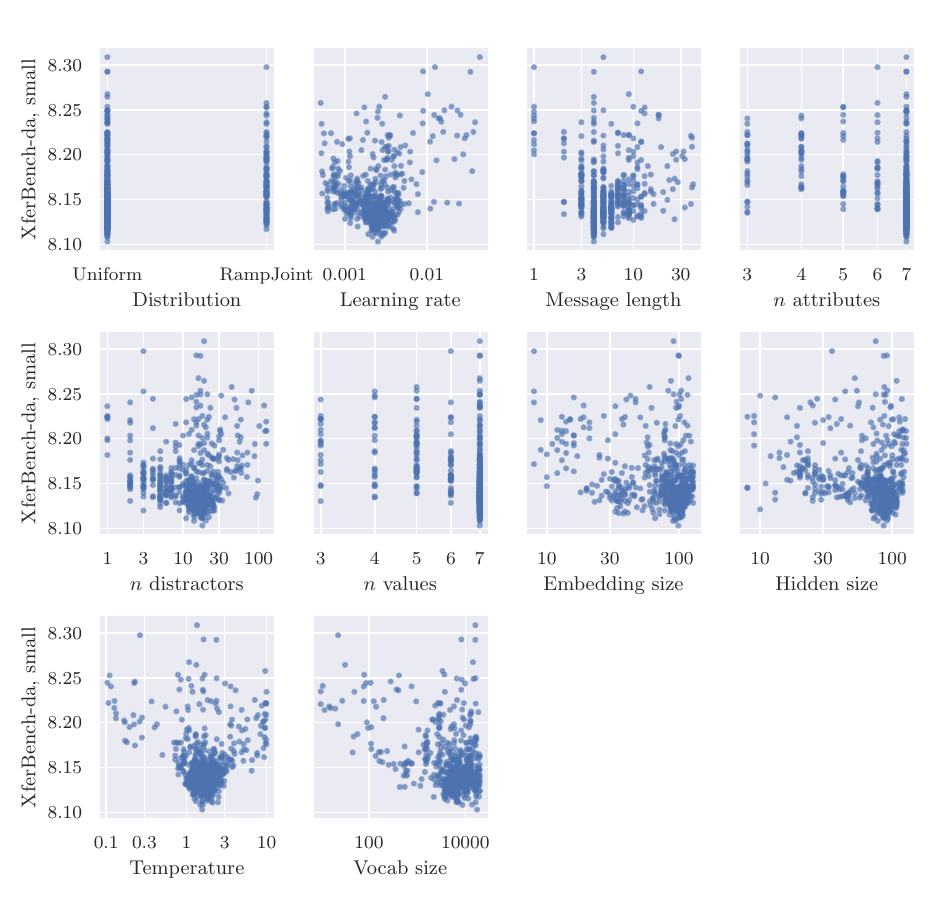}{Objective values for Search 1 by individual hyperparameter.}
\slicefig{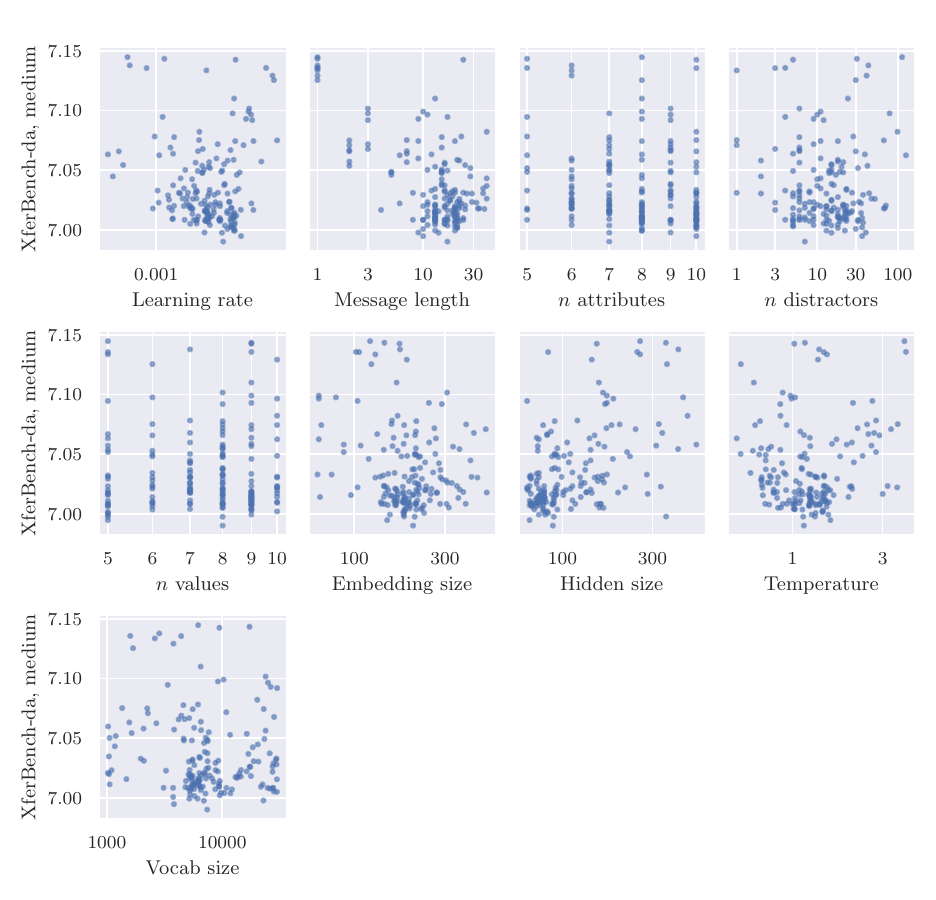}{Objective values for Search 2 by individual hyperparameter.}
\slicefig{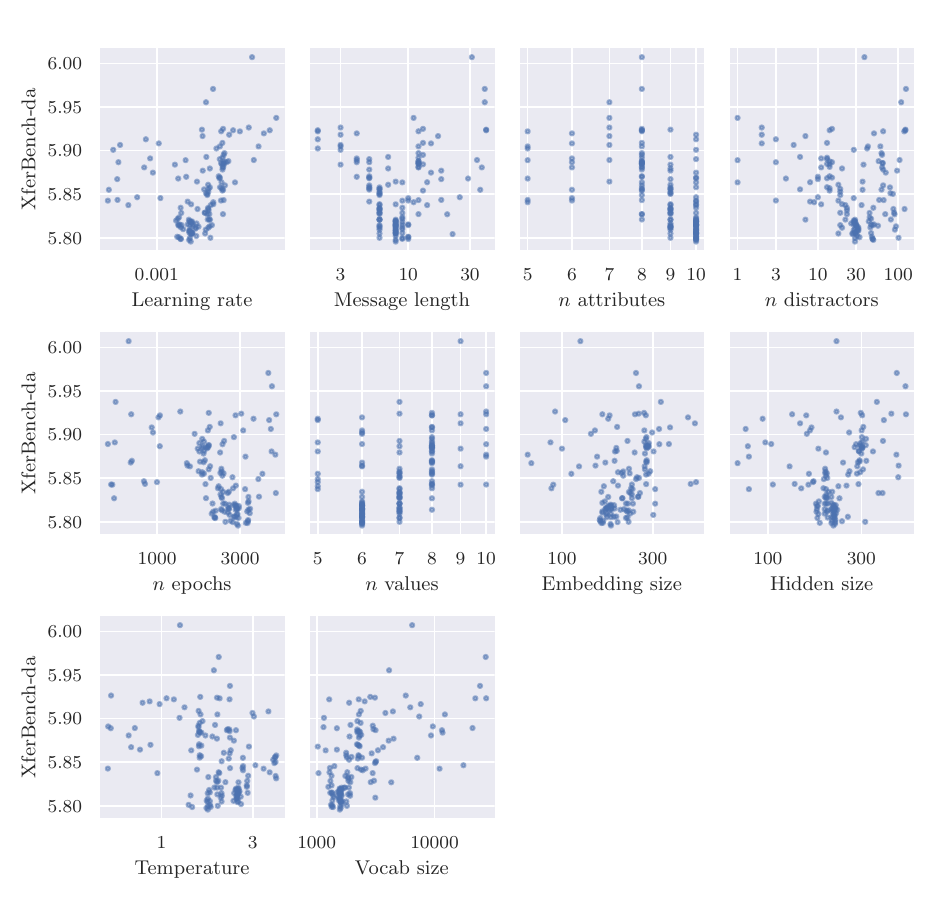}{Objective values for Search 3 by individual hyperparameter.}
\slicefig{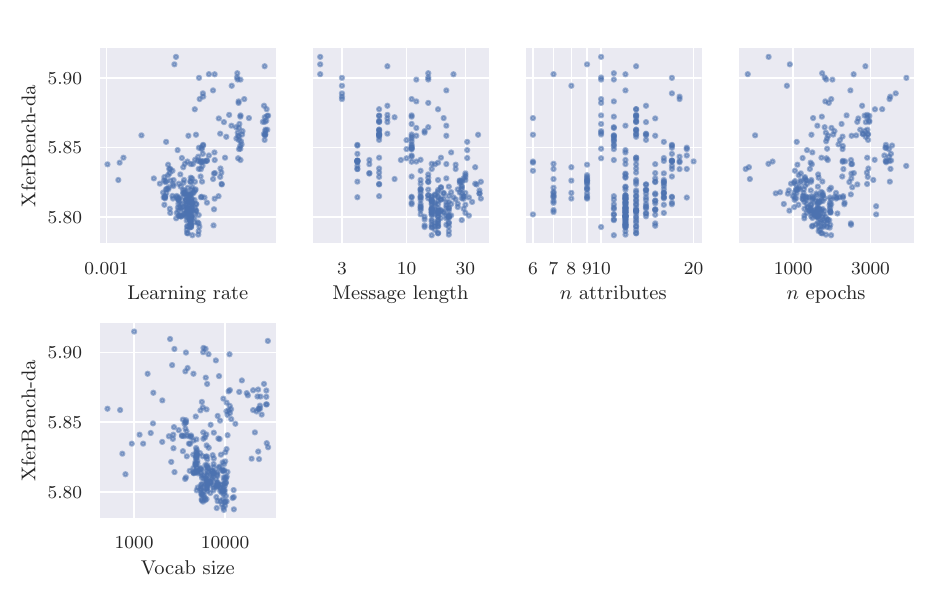}{Objective values for Search 4 by individual hyperparameter.}

\section{Varying Neural Architecture}%
\label{sec:arch-scatter}
\slicefig{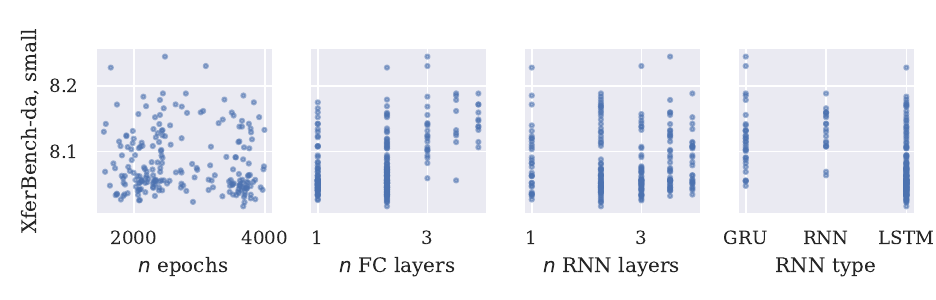}{Objective values for Search 7 by individual hyperparameter.}

In a follow-up experiment we test different neural architectures for the sender and receiver agents.
In particular, we test different numbers of fully connected layers ($\{1,\dots,5\}$), RNN layers ($\{1,\dots,5\}$), and RNN types (GRU, LSTM, Elman) \citep{elman,lstm,gru}.
The number of epochs was also allowed to vary in the event that increasing the number of parameters benefited from longer training.
\Cref{fig:slice-4} displays the results of this experiment.

The fully connected layers (which surround the sender's and receiver's RNN) have the same hidden size and are separated by tanh activations.
The RNN layers vary according to a standard stacked architecture.
The \emph{RNN} cell type refers to a plain Elman RNN\@.
The small variant of XferBench-da was used for the objective.

From \cref{fig:slice-4}, we see that LSTMs outperform GRUs (used for the main experiments) and RNNs by a large margin.
On the other hand, Using $2$ instead of $1$ layer (used for the main experiments) provides a smaller performance gain on XferBench-d while further increasing the layers does not show improvement.
The number of epochs did not have a notable effect.


\typeout{INFO: \arabic{comment} comments.}
\end{document}